\documentclass{article} 
\usepackage{acl2019,times}

\usepackage{amsmath,amsfonts,bm}

\def\eqref#1{equation~\ref{#1}}

\def\1{\bm{1}}

\DeclareMathAlphabet{\mathsfit}{\encodingdefault}{\sfdefault}{m}{sl}
\SetMathAlphabet{\mathsfit}{bold}{\encodingdefault}{\sfdefault}{bx}{n}

\usepackage{hyperref}
\usepackage{url}
\usepackage{tikz-dependency}
\usepackage{subcaption}
\usepackage[space]{grffile}
\usepackage{hhline}
\usepackage{booktabs}

\aclfinalcopy
\title{BART: Denoising Sequence-to-Sequence Pre-training for Natural Language Generation, Translation, and Comprehension }

\author{Mike Lewis*, Yinhan Liu*, Naman Goyal*,
Marjan Ghazvininejad,\\ \textbf{Abdelrahman Mohamed, Omer Levy, Ves Stoyanov, Luke Zettlemoyer } \\
Facebook AI\\
\texttt{\{mikelewis,yinhanliu,naman\}@fb.com} \\
}

\begin{document}

\maketitle
\begin{abstract}

We present BART, a denoising autoencoder for pretraining sequence-to-sequence models. BART is trained by (1) corrupting text with an arbitrary noising function, and (2) learning a model to reconstruct the original text. It uses a standard Tranformer-based neural machine translation architecture which, despite its simplicity, can be seen as generalizing BERT (due to the bidirectional encoder), GPT (with the left-to-right decoder), and many other more recent pretraining schemes. We evaluate a number of noising approaches, finding the best performance by both randomly shuffling the order of the original sentences and using a novel in-filling scheme, where spans of text are replaced with a single mask token. BART is particularly effective when fine tuned for text generation but also works well for comprehension tasks. It matches the performance of RoBERTa with comparable training resources on GLUE and SQuAD, achieves new state-of-the-art results on a range of abstractive dialogue, question answering, and summarization tasks, with gains of up to 6 ROUGE. BART also provides a 1.1 BLEU increase over a back-translation system for machine translation, with only target language pretraining. We also report ablation experiments that replicate other pretraining schemes within the BART framework, to better measure which factors most influence end-task performance. 

\end{abstract}

\section{Introduction}
Self-supervised methods have achieved remarkable success in a wide range of NLP tasks~ \cite{word2vec,elmo,bert,spanbert,xlnet,roberta}. 
The most successful approaches have been variants of masked language models, which are denoising autoencoders that are trained to reconstruct text where a random subset of the words has been masked out. Recent work has shown gains by improving the distribution of masked tokens~\cite{spanbert}, the order in which masked tokens are predicted~\cite{xlnet}, and the available context for replacing masked tokens~\cite{unilm}. However, these methods typically focus on particular types of end tasks (e.g. span prediction, generation, etc.), limiting their applicability. 

\begin{figure*}[t!]
    \centering
    \begin{subfigure}[b]{0.45\textwidth}
        \centering
        \includegraphics[height=1.2in]{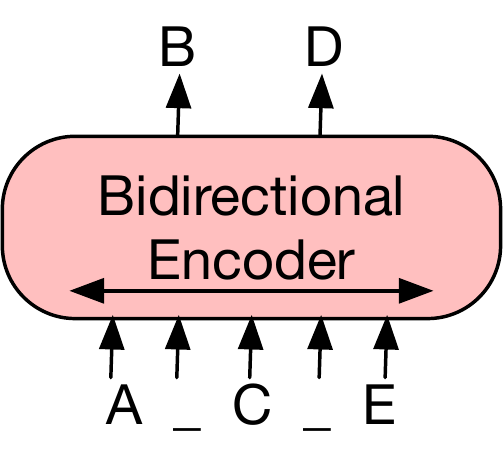}
        \caption{BERT: Random tokens are replaced with masks, and the document is encoded bidirectionally. Missing tokens are predicted independently, so BERT cannot easily be used for generation.}
    \end{subfigure}~~~
    \begin{subfigure}[b]{0.45\textwidth}
        \centering
        \includegraphics[height=1.2in]{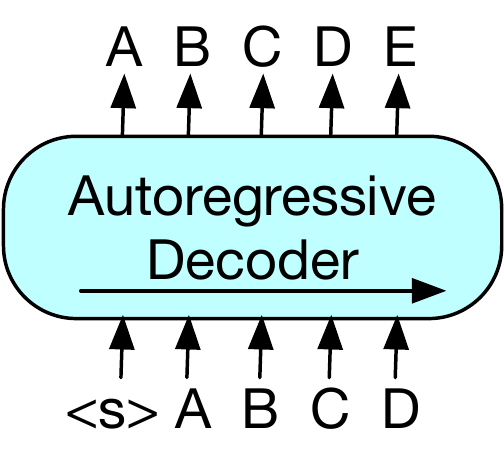}
        \caption{GPT: Tokens are predicted auto-regressively, meaning GPT can be used for generation. However words can only condition on leftward context, so it cannot learn bidirectional interactions.}
    \end{subfigure}
    \\
    \begin{subfigure}[b]{\textwidth}
        \centering
        \includegraphics[height=1.2in]{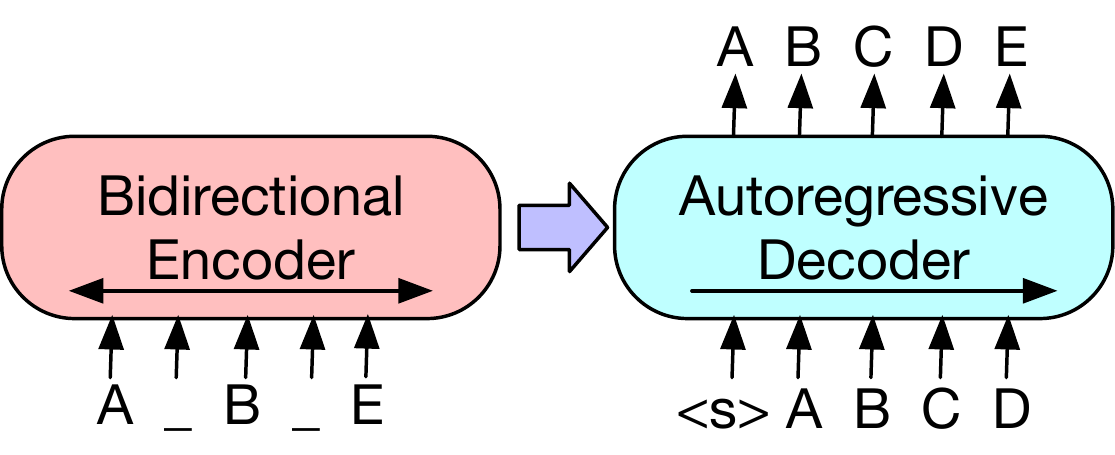}
        \caption{BART: Inputs to the encoder need not be aligned with decoder outputs, allowing arbitary noise transformations. Here, a document has been corrupted by replacing spans of text with mask symbols. The corrupted document (left) is encoded with a bidirectional model, and then the likelihood of the original document (right) is calculated with an autoregressive decoder. For fine-tuning, an uncorrupted document is input to both the encoder and decoder, and we use representations from the final hidden state of the decoder.}
    \end{subfigure}
    \caption{A schematic comparison of BART with BERT \cite{bert} and GPT \cite{gpt}.}
    \label{figure:summary}
\end{figure*}

In this paper, we present BART, which pre-trains a model combining Bidirectional and Auto-Regressive Transformers. BART is a denoising autoencoder built with a sequence-to-sequence model that is applicable to a very wide range of end tasks. Pretraining has two stages (1) text is corrupted with an arbitrary noising function, and (2) a sequence-to-sequence model is learned to reconstruct the original text. BART uses a standard Tranformer-based neural machine translation architecture which, despite its simplicity, can be seen as generalizing BERT (due to the bidirectional encoder), GPT (with the left-to-right decoder), and many other more recent pretraining schemes (see Figure \ref{figure:summary}). 

A key advantage of this setup is the noising flexibility; arbitrary transformations can be applied to the original text, including changing its length. We evaluate a number of noising approaches, finding the best performance by both randomly shuffling the order of the original sentences and using a novel in-filling scheme, where arbitrary length spans of text (including zero length) are replaced with a single mask token. This approach generalizes the original word masking and next sentence prediction objectives in BERT by forcing the model to reason more about overall sentence length and make longer range transformations to the input.

BART is particularly effective when fine tuned for text generation but also works well for comprehension tasks. It matches the performance of RoBERTa \cite{roberta} with comparable training resources on GLUE \cite{glue} and SQuAD \cite{squad}, and achieves new state-of-the-art results on a range of abstractive dialogue, question answering, and summarization tasks. For example, it improves performance by 6 ROUGE over previous work on XSum \cite{xsum}. 

BART also opens up new ways of thinking about fine tuning. We present a new scheme for machine translation where a BART model is stacked above a few additional transformer layers. These layers are trained to essentially translate the foreign language to noised English, by propagation through BART, thereby using BART as a pre-trained target-side language model. This approach improves performance over a strong back-translation MT baseline by 1.1 BLEU on the WMT Romanian-English benchmark.

To better understand these effects, we also report an ablation analysis that replicates other recently proposed training objectives. This study allows us to carefully control for a number of factors, including data and optimization parameters, which have been shown to be as important for overall performance as the selection of training objectives~\cite{roberta}. We find that BART exhibits the most consistently strong performance across the full range of tasks we consider. 


    \begin{figure*}[t]
        \centering
        \includegraphics[height=1.2in]{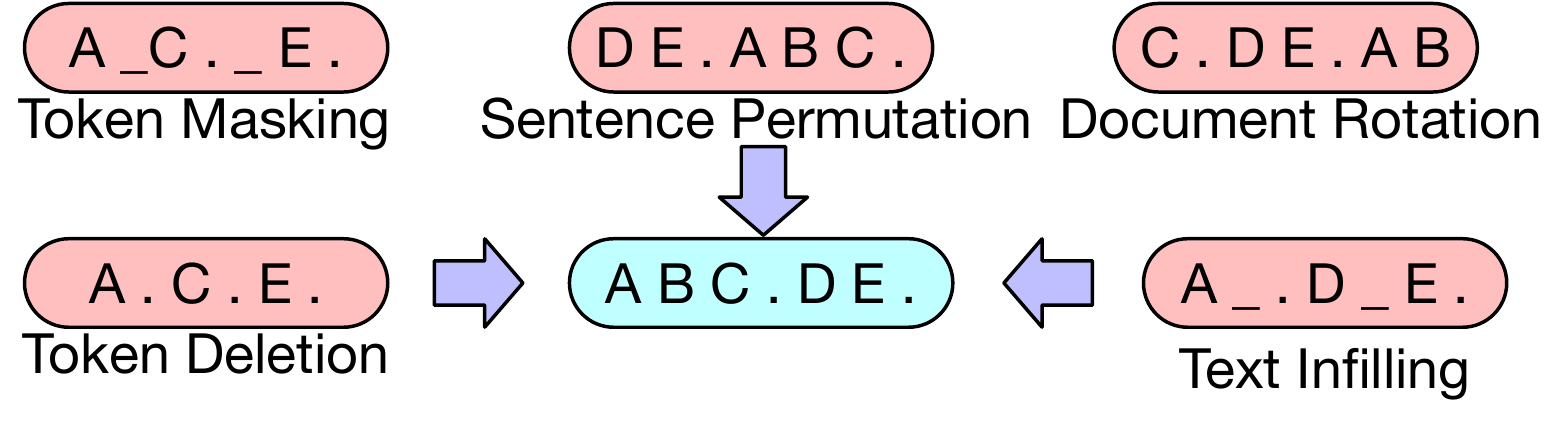}
        \caption{Transformations for noising the input that we experiment with. These transformations can be composed. \label{figure:noise}}
    \end{figure*}%

\section{Model}
BART is a denoising autoencoder that maps a corrupted document to the original document it was derived from. It is implemented as a sequence-to-sequence model with a bidirectional encoder over corrupted text and a left-to-right autoregressive decoder. For pre-training, we optimize the negative log likelihood of the original document.

\subsection{Architecture}
BART uses the standard sequence-to-sequence Transformer architecture from \cite{vaswani:2017}, except, following GPT, that we modify ReLU activation functions to GeLUs \citep{gelu} and initialise parameters from $\mathcal{N}(0, 0.02)$. For our base model, we use 6 layers in the encoder and decoder, and for our large model we use 12 layers in each. The architecture is closely related to that used in BERT, with the following differences: (1) each layer of the decoder additionally performs cross-attention over the final hidden layer of the encoder (as in the transformer sequence-to-sequence model); and (2) BERT uses an additional feed-forward network before word-prediction, which BART does not. In total, BART contains roughly 10\% more parameters than the equivalently sized BERT model. 

\subsection{Pre-training BART}
BART is trained by corrupting documents and then optimizing a reconstruction loss---the cross-entropy between the decoder's output and the original document. Unlike existing denoising autoencoders, which are tailored to specific noising schemes, BART allows us to apply \emph{any} type of document corruption.
In the extreme case, where all information about the source is lost, BART is equivalent to a language model.

We experiment with several previously proposed and novel transformations, but we believe there is a significant potential for development of other new alternatives. The transformations we used are summarized below, and examples are shown in Figure~\ref{figure:noise}.

\paragraph{Token Masking} Following BERT \cite{bert}, random tokens are sampled and replaced with \texttt{[MASK]} elements.

\paragraph{Token Deletion} Random tokens are deleted from the input. In contrast to token masking, the model must decide which positions are missing inputs.

\paragraph{Text Infilling} A number of text spans are sampled, with span lengths drawn from a Poisson distribution ($\lambda=3$). Each span is replaced with a \emph{single} \texttt{[MASK]} token. 0-length spans correspond to the insertion of \texttt{[MASK]} tokens.
Text infilling is inspired by SpanBERT \cite{spanbert}, but SpanBERT samples span lengths from a different  (clamped geometric) distribution, and replaces each span with a sequence of \texttt{[MASK]} tokens of exactly the same length. Text infilling teaches the model to predict how many tokens are missing from a span.

\paragraph{Sentence Permutation} A document is divided into sentences based on full stops, and these sentences are shuffled in a random order.

\paragraph{Document Rotation} A token is chosen uniformly at random, and the document is rotated so that it begins with that token. This task trains the model to identify the start of the document.


\begin{figure*}[t!]
    \centering
    \begin{subfigure}[b]{0.45\textwidth}
        \centering
        \includegraphics[height=1.2in]{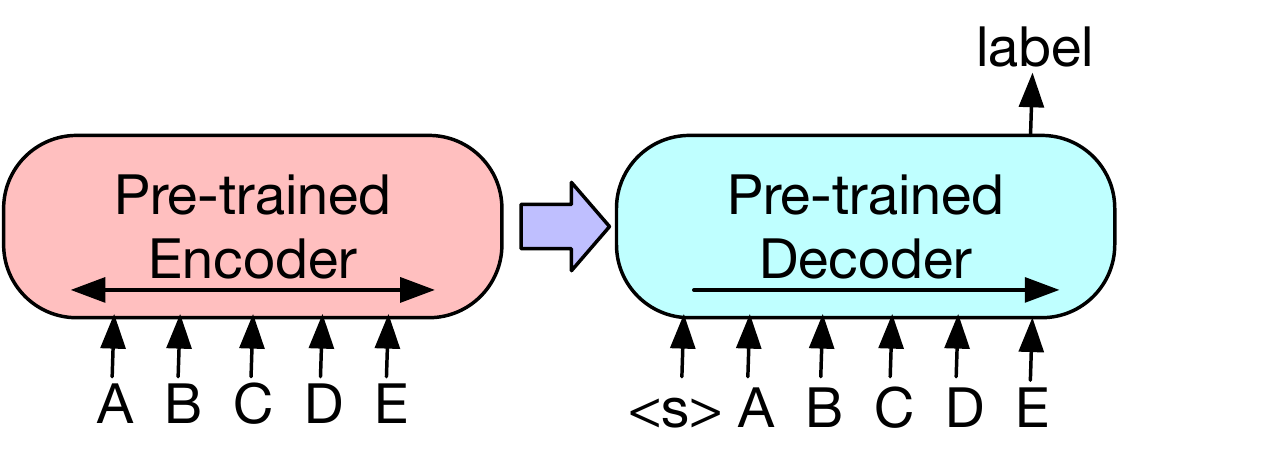}
        \caption{\label{figure:finetune_classification} To use BART for classification problems, the same input is fed into the encoder and decoder, and the representation from the final output is used.}
    \end{subfigure}~~~
    \begin{subfigure}[b]{0.45\textwidth}
        \centering
        \includegraphics[height=1.2in]{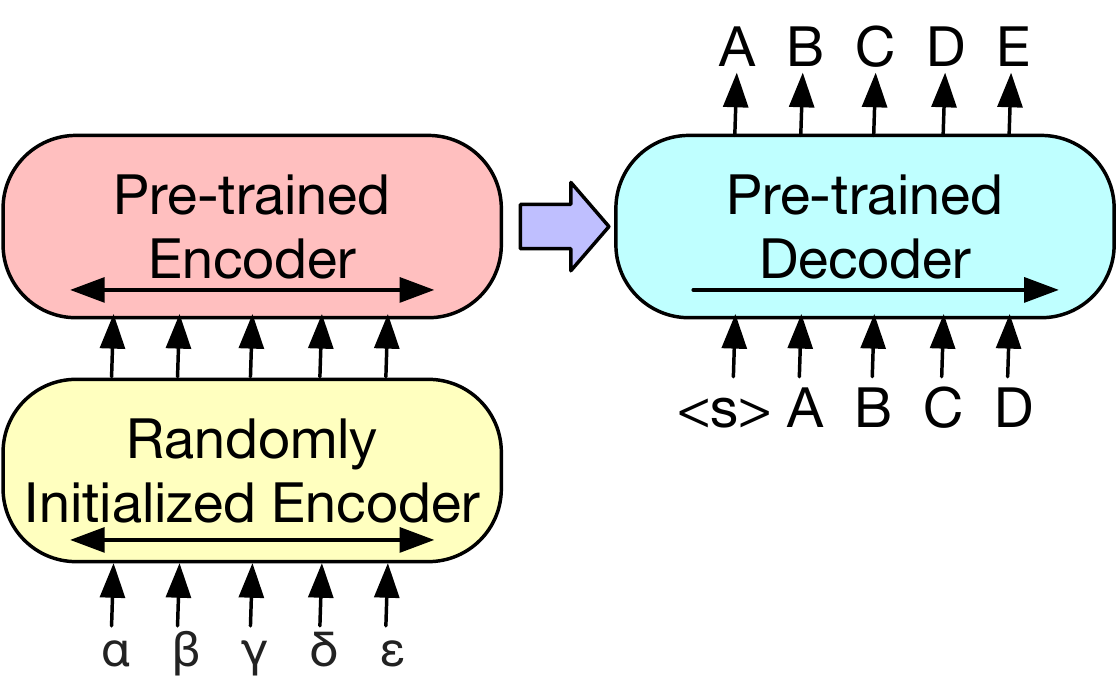}
        \caption{\label{figure:finetune_translation} For machine translation, we learn a small additional encoder that replaces the word embeddings in BART. The new encoder can use a disjoint vocabulary.
        \label{figure:MT}}
    \end{subfigure}
\caption{Fine tuning BART for classification and translation.}
\end{figure*}

\section{Fine-tuning BART}


The representations produced by BART can be used in several ways for downstream applications.

\subsection{Sequence Classification Tasks}
For sequence classification tasks, the same input is fed into the encoder and decoder, and the final hidden state of the final decoder token is fed into new multi-class linear classifier.  This approach is related to the CLS token in BERT; however we add the additional token to the \emph{end} so that representation for the token in the decoder can attend to decoder states from the complete input (Figure \ref{figure:finetune_classification}).

\subsection{Token Classification Tasks}
For token classification tasks, such as answer endpoint classification for SQuAD, we feed the complete document into the encoder and decoder, and use the top hidden state of the decoder as a representation for each word. This representation is used to classify the token.

\subsection{Sequence Generation Tasks}
Because BART has an autoregressive decoder, it can be directly fine tuned for sequence generation tasks such as abstractive question answering and summarization. In both of these tasks, information is copied from the input but manipulated, which is closely related to the denoising pre-training objective. Here, the encoder input is the input sequence, and the decoder generates outputs autoregressively.


\subsection{Machine Translation}
\label{sec:tuning-MT}
We also explore using BART to improve machine translation decoders for translating into English. Previous work \newcite{edunov2019pre} has shown that models can be improved by incorporating pre-trained encoders, but gains from using pre-trained language models in decoders have been limited. We show that it is possible to use the entire BART model (both encoder and decoder) as a single pretrained decoder for machine translation, by adding a new set of encoder parameters that are learned from bitext (see Figure \ref{figure:MT}).

More precisely, we replace BART's encoder embedding layer with a new randomly initialized encoder. The model is trained end-to-end, which trains the new encoder to map foreign words into an input that BART can de-noise to English. The new encoder can use a separate vocabulary from the original BART model.



We train the source encoder in two steps, in both cases backpropagating the cross-entropy loss from the output of the BART model. In the first step, we freeze most of BART parameters and only update the randomly initialized source encoder, the BART positional embeddings, and the self-attention input projection matrix of BART's encoder first layer.
In the second step, we train all model parameters for a small number of iterations. 

\section{Comparing Pre-training Objectives}
\label{section:ablations}
BART supports a much wider range of noising schemes during pre-training than previous work. We compare a range of options using base-size models (6 encoder and 6 decoder layers, with a hidden size of 768), evaluated on a representative subset of the tasks we will consider for the full large scale experiments in \S\ref{sec:large-scale}.

\subsection{Comparison Objectives}
\label{sec:baselines}
While many pre-training objectives have been proposed, fair comparisons between these have been difficult to perform, at least in part due to differences in training data, training resources, architectural differences between models, and fine-tuning procedures. 
We re-implement strong pre-training approaches recently proposed for discriminative and generation tasks.
We aim, as much as possible, to control for differences unrelated to the pre-training objective.
However, we do make minor changes to the learning rate and usage of layer normalisation in order to improve performance (tuning these separately for each objective). 
For reference, we compare our implementations with published numbers from BERT, which was also trained for 1M steps on a combination of books and Wikipedia data.
We compare the following approaches:

\begin{table*}[ht]
\begin{center}
\begin{tabular}{lccccccccc}
\toprule
\textbf{Model} & \textbf{SQuAD 1.1}     & \textbf{MNLI}  & \textbf{ELI5} & \textbf{XSum} & \textbf{ConvAI2}  & \textbf{CNN/DM}  \\
                                & F1                         & Acc          &    PPL        & PPL           &        PPL                 & PPL \\
\midrule
BERT Base \cite{bert}           &        88.5          & \textbf{84.3}    &      -        &       -       &            -         & -      \\
\midrule
Masked Language Model           &        90.0     &         83.5      &   24.77    &   ~~7.87     & 12.59                      &~~7.06\\ 
Masked Seq2seq                  &        87.0               &        82.1       &     23.40     &~~6.80         & 11.43                     & ~~6.19\\ 
Language Model                  &        76.7              &        80.1       &\textbf{21.40} &~~7.00         & 11.51    &                  ~~6.56\\ 
Permuted Language Model         &        89.1                &      83.7       &     24.03     &     ~~7.69   & 12.23                      &~~6.96\\ 
Multitask Masked Language Model &        89.2              &        82.4       &      23.73   &     ~~7.50    &   12.39                   &~~6.74\\ 
\midrule
BART Base             &   \\
w/ Token Masking                 &      90.4                &      84.1        &     25.05     &~~7.08          & 11.73                 &~~6.10\\ 
w/ Token Deletion                &       90.4                &      84.1        &     24.61     &~~6.90          & 11.46                 &~~5.87\\ 
w/ Text Infilling                &       \textbf{90.8}                &     84.0        &     24.26     &~~\textbf{6.61} & \textbf{11.05}                &~~5.83\\ 
w/ Document Rotation             &       77.2                &      75.3        &     53.69     &17.14          & 19.87                &10.59\\ 
w/ Sentence Shuffling            &       85.4                &      81.5        &     41.87     &10.93          & 16.67               &~~7.89\\ 
w/ Text Infilling + Sentence Shuffling & \textbf{90.8}                &      83.8        &     24.17     &~~6.62          & 11.12               &~~\textbf{5.41} \\ 
\bottomrule
\end{tabular}
\end{center}

\caption{\label{table:ablations} Comparison of pre-training objectives. All models are of comparable size and are trained for 1M steps on a combination of books and Wikipedia data. Entries in the bottom two blocks are trained on identical data using the same code-base, and fine-tuned with the same procedures. Entries in the second block are inspired by pre-training objectives proposed in previous work, but have been simplified to focus on evaluation objectives (see \S\ref{sec:baselines}). Performance varies considerably across tasks, but the BART models with text infilling demonstrate the most consistently strong performance.} 
\end{table*}

\paragraph{Language Model} Similarly to GPT \citep{gpt}, we train a left-to-right Transformer language model. This model is equivalent to the BART decoder, without cross-attention.

\paragraph{Permuted Language Model} Based on XLNet \citep{xlnet}, we sample 1/6 of the tokens, and generate them in a random order autoregressively. For consistency with other models, we do not implement the relative positional embeddings or attention across segments from XLNet.

\paragraph{Masked Language Model} Following BERT \citep{bert}, we replace 15\% of tokens with [MASK] symbols, and train the model to independently predict the original tokens. 

\paragraph{Multitask Masked Language Model} As in UniLM \citep{unilm}, we train a Masked Language Model with additional self-attention masks. Self attention masks are chosen randomly in with the follow proportions: 1/6 left-to-right, 1/6 right-to-left, 1/3 unmasked, and 1/3 with the first 50\% of tokens unmasked and a left-to-right mask for the remainder.

\paragraph{Masked Seq-to-Seq} Inspired by MASS \citep{mass}, we mask a span containing 50\% of tokens, and train a sequence to sequence model to predict the masked tokens.
\vspace{10pt}

 

  
  

For the Permuted LM, Masked LM and Multitask Masked LM, we use two-stream attention \citep{xlnet} to efficiently compute likelihoods of the output part of the sequence (using a diagonal self-attention mask on the output to predict words left-to-right). 

We experiment with (1) treating the task as a standard sequence-to-sequence problem, where the source input to the encoder and the target is the decoder output
, or (2) adding the source as prefix to the target in the decoder, with a loss only on the target part of the sequence. 
We find the former works better for BART models, and the latter for other models.

To most directly compare our models on their ability to model their fine-tuning objective (the log likelihood of the human text), we report perplexity in Table \ref{table:ablations}.



\subsection{Tasks}
\paragraph{SQuAD} \cite{squad}a an extractive question answering task on Wikipedia paragraphs. Answers are text spans extracted from a given document context. Similar to BERT \cite{bert}, we use concatenated question and context as input to the encoder of BART, and additionally pass them to the decoder. The model includes classifiers to predict the start and end indices of each token. 
\paragraph{MNLI} \citep{mnli}, a bitext classification task to predict whether one sentence entails another. The fine-tuned model concatenates the two sentences with  appended an EOS token, and passes them to both the BART encoder and decoder. In contrast to BERT, the representation of the EOS token is used to classify the sentences relations. 
\paragraph{ELI5} \citep{eli5}, a long-form abstractive question answering dataset. Models generate answers conditioned on the concatenation of a question and supporting documents. 
\paragraph{XSum} \citep{xsum}, a news summarization dataset with highly abstractive summaries. 
\paragraph{ConvAI2} \citep{convai2}, a dialogue response generation task, conditioned on context and a persona.
\paragraph{CNN/DM} \citep{cnn}, a news summarization dataset. Summaries here are typically closely related to source sentences. 

\subsection{Results}
Results are shown in Table \ref{table:ablations}. Several trends are clear:

\paragraph{Performance of pre-training methods varies significantly across tasks} The effectiveness of pre-training methods is highly dependent on the task. For example, a simple language model achieves the best ELI5 performance, but the worst SQUAD results.

\paragraph{Token masking is crucial} Pre-training objectives based on rotating documents or permuting sentences perform poorly in isolation. The successful methods either use token deletion or masking, or self-attention masks. Deletion appears to outperform masking on generation tasks.

\paragraph{Left-to-right pre-training improves generation} The Masked Language Model and the Permuted Language Model perform less well than others on generation, and are the only models we consider that do not include left-to-right auto-regressive language modelling during pre-training.

\paragraph{Bidirectional encoders are crucial for SQuAD} As noted in previous work \cite{bert}, just left-to-right decoder performs poorly on SQuAD, because future context is crucial in classification decisions. However, BART achieves similar performance with only half the number of bidirectional layers.

\paragraph{The pre-training objective is not the only important factor} Our Permuted Language Model performs less well than XLNet \cite{xlnet}. Some of this difference is likely due to not including other architectural improvements, such as relative-position embeddings or segment-level recurrence. 

\paragraph{Pure language models perform best on ELI5} The ELI5 dataset is an outlier, with much higher perplexities than other tasks, and is the only generation task where other models outperform BART. A pure language model performs best, suggesting that BART is less effective when the output is only loosely constrained by the input.

\paragraph{BART achieves the most consistently strong performance.} With the exception of ELI5, BART models using text-infilling perform well on all tasks.

\begin{table*}[h!t]
\centering
\begin{tabular}{lcccccccccc}
\toprule
        & \textbf{SQuAD 1.1} & \textbf{SQuAD 2.0} & \textbf{MNLI} & \textbf{SST} & \textbf{QQP}	& \textbf{QNLI} &  \textbf{STS-B}& \textbf{RTE}&  \textbf{MRPC} & \textbf{CoLA}\\
        &   EM/F1  &  EM/F1    & m/mm &Acc &Acc& Acc
        &Acc&Acc&Acc&Mcc\\
\midrule
BERT  & 84.1/90.9 &  79.0/81.8  & 86.6/-  & 93.2      & 91.3&92.3&90.0&70.4&88.0&60.6   \\
UniLM   &     -/-    &    80.5/83.4 &  87.0/85.9  &  94.5    &  - & 92.7 & -&70.9&-&61.1 \\
XLNet   &   \textbf{89.0}/94.5  &86.1/88.8  &  89.8/- &  95.6    &  91.8&93.9&91.8 &83.8&89.2 &63.6  \\
RoBERTa &  88.9/\textbf{94.6} &  \textbf{86.5/89.4}  &  \textbf{90.2/90.2}& 96.4     &  92.2	&94.7	&\textbf{92.4}	&86.6&	\textbf{90.9}&	\textbf{68.0}
   \\
BART    &  88.8/\textbf{94.6} & 86.1/89.2  &  89.9/90.1   &   \textbf{96.6} 
&  \textbf{92.5}	&\textbf{94.9}&	91.2&	\textbf{87.0}&	90.4	&62.8
\\
\bottomrule
\end{tabular}
\caption{\label{table:discriminative}Results for large models on SQuAD and GLUE tasks. BART performs comparably to RoBERTa and XLNet, suggesting that BART's uni-directional decoder layers do not reduce performance on discriminative tasks.
}
\end{table*}

\begin{table*}[t]
\centering
\begin{tabular}{lccccccccc}
\toprule
                          & \multicolumn{3}{c}{\textbf{CNN/DailyMail}} &  \multicolumn{3}{c}{\textbf{XSum}} \\
                          & R1        & R2        & RL           & R1     & R2     & RL     \\
\midrule
Lead-3                    & 40.42     & 17.62     & 36.67            & 16.30  & 1.60   & 11.95 \\
PTGEN \cite{see:2017}     & 36.44     & 15.66     & 33.42            & 29.70  & 9.21   & 23.24  \\
PTGEN+COV \cite{see:2017} & 39.53     & 17.28     & 36.38            & 28.10  & 8.02   & 21.72  \\
UniLM                     &  43.33    &  20.21    & 40.51            &   -    &    -   &    -   \\
BERTSUMABS \citep{bertsum}               & 41.72     & 19.39     & 38.76            &    38.76 & 16.33 & 31.15  \\
BERTSUMEXTABS  \citep{bertsum}            & 42.13     & 19.60     & 39.18            & 38.81  & 16.50  & 31.27   \\
\midrule
BART                      & \textbf{44.16}      & \textbf{21.28}      & \textbf{40.90}         & \textbf{45.14}   & \textbf{22.27}   & \textbf{37.25} \\
\bottomrule
\end{tabular}
\caption{Results on two standard summarization datasets. BART outperforms previous work on summarization on two tasks and all metrics, with gains of roughly 6 points on the more abstractive dataset.}
\end{table*}

\section{Large-scale Pre-training Experiments}
\label{sec:large-scale}

Recent work has shown that downstream performance can dramatically improve when pre-training is scaled to large batch sizes \citep{xlnet,roberta} and corpora. To test how well BART performs in this regime, and to create a useful model for downstream tasks, we trained BART using the same scale as the RoBERTa model.

\subsection{Experimental Setup}
We pre-train a large model with 12 layers in each of the encoder and decoder, and a hidden size of 1024. Following RoBERTa \cite{roberta}, we use a batch size of 8000, and train the model for 500000 steps. Documents are tokenized with the same byte-pair encoding as GPT-2 \cite{gpt2}. Based on the results in Section \S\ref{section:ablations}, we use a combination of text infilling and sentence permutation. We mask 30\% of tokens in each document, and permute all sentences. 
Although sentence permutation only shows significant additive gains on the CNN/DM summarization dataset, we hypothesised that larger pre-trained models may be better able to learn from this task.
To help the model better fit the data, we disabled dropout for the final 10\% of training steps. 
We use the same pre-training data as \citet{roberta}, consisting of 160Gb of news, books, stories, and web text.

\subsection{Discriminative Tasks}
Table \ref{table:discriminative} compares the performance of BART with several recent approaches on the well-studied SQuAD and GLUE tasks \citep{warstadt2018neural,socher2013recursive,dolan2005automatically,agirre2007semantic,williams2018broad,dagan2006pascal,levesque2011winograd}. 

The most directly comparable baseline is RoBERTa, which was pre-trained with the same resources, but a different objective. Overall, BART performs similarly, with only small differences between the models on most tasks. suggesting that BART's improvements on generation tasks do not come at the expense of classification performance.

\subsection{Generation Tasks}
We also experiment with several text generation tasks. BART is fine-tuned as a standard sequence-to-sequence model from the input to the output text.
During fine-tuning we use a label smoothed cross entropy loss \citep{labelsmoothing}, with the smoothing parameter set to 0.1. During generation, we set beam size as 5, remove duplicated trigrams in beam search, 
and tuned the model with min-len, max-len, length penalty on the validation set \cite{summarization_hacks}.


\paragraph{Summarization}
To provide a comparison with the state-of-the-art in summarization, we present results on two summarization datasets, CNN/DailyMail and XSum, which have distinct properties. 

Summaries in the CNN/DailyMail tend to resemble source sentences. Extractive models do well here, and even the baseline of the first-three source sentences is highly competitive. Nevertheless, BART outperforms all existing work.

In contrast, XSum is highly abstractive, and extractive models perform poorly. BART outperforms the best previous work, which leverages BERT, by roughly 6.0 points on all ROUGE metrics---representing a significant advance in performance on this problem. Qualitatively, sample quality is high (see \S\ref{section:qualitative}). 

\begin{table}[t]
\centering
\begin{tabular}{lccc}
\toprule
                     & \multicolumn{2}{c}{\textbf{ConvAI2}} & 
                     \\
                     & Valid F1           & Valid PPL         \\
\midrule
Seq2Seq + Attention	&   16.02            &       35.07 \\             
Best System         &   19.09            &       17.51        \\
BART                 &  \textbf{20.72}            &     \textbf{11.85}           &         \\
\bottomrule
\end{tabular}
\caption{BART outperforms previous work on conversational response generation. Perplexities are renormalized based on official tokenizer for ConvAI2. 
}
\end{table}
\begin{table}[t]
\centering
\begin{tabular}{lcccc}
\toprule
                     & \multicolumn{3}{c}{\textbf{ELI5}}  \\
                     & R1           & R2           & RL                \\ 
\midrule
Best Extractive      &   23.5           &   3.1           &   17.5                 \\
Language Model       &    27.8          &     4.7         &    23.1                \\
Seq2Seq              &    28.3          &     5.1         &   22.8                 \\
Seq2Seq Multitask    &     28.9         &        5.4      &     23.1               \\
BART                 &        \textbf{30.6}      &    \textbf{ 6.2  }       &    \textbf{24.3 }      \\
\bottomrule
\end{tabular}
\caption{BART achieves state-of-the-art results on the challenging ELI5 abstractive question answering dataset. Comparison models are from \citet{eli5}.}
\end{table}
\paragraph{Dialogue} We evaluate dialogue response generation on \textsc{ConvAI2} \citep{convai2}, in which agents must generate responses conditioned on both the previous context and a textually-specified persona. BART outperforms previous work on two automated metrics.

\paragraph{Abstractive QA}
We use the recently proposed ELI5 dataset to test the model's ability to generate long free-form answers. We find BART outperforms the best previous work by 1.2 ROUGE-L, but the dataset remains a challenging, because answers are only weakly specified by the question.


\subsection{Translation}



\begin{table}[t]
\centering
\begin{tabular}{lcc}
\toprule
&  RO-EN	 \\
\midrule
Baseline& 36.80\\
Fixed BART & 36.29\\
Tuned BART & \textbf{37.96}\\
\bottomrule
\end{tabular}
\caption{The performance (BLEU) of baseline and BART on WMT'16 RO-EN augmented with back-translation data. BART improves over a strong back-translation (BT) baseline by using monolingual English pre-training.}
\label{tab:MT_results}
\end{table}
\begin{table*}[ht]
\centering
\begin{tabular}{p{9cm}p{6cm}}
\toprule
\textbf{Source Document (abbreviated)}   & \textbf{BART Summary} \\
\midrule
The researchers examined three types of coral in reefs off the coast of Fiji ... The researchers found when fish were plentiful, they would eat algae and seaweed off the corals, which appeared to leave them more resistant to the bacterium Vibrio coralliilyticus, a bacterium associated with bleaching. The researchers suggested the algae, like warming temperatures, might render the corals' chemical defenses less effective, and the fish were protecting the coral by removing the algae. & Fisheries off the coast of Fiji are protecting coral reefs from the effects of global warming, according to a study in the journal Science.\\
\midrule
Sacoolas, who has immunity as a diplomat's wife, was involved in a traffic collision ... Prime Minister Johnson was questioned about the case while speaking to the press at a hospital in Watford. He said, ``I hope that Anne Sacoolas will come back ... if we can't resolve it then of course I will be raising it myself personally with the White House.'' & Boris Johnson has said he will raise the issue of US diplomat Anne Sacoolas' diplomatic immunity with the White House. \\
\midrule
According to Syrian state media, government forces began deploying into previously SDF controlled territory yesterday. ... On October 6, US President Donald Trump and Turkish President Recep Tayyip Erdoğan spoke on the phone. Then both nations issued statements speaking of an imminent incursion into northeast Syria ... . On Wednesday, Turkey began a military offensive with airstrikes followed by a ground invasion. & Syrian government forces have entered territory held by the US-backed Syrian Democratic Forces (SDF) in response to Turkey's incursion into the region. \\
\midrule
This is the first time anyone has been recorded to run a full marathon of 42.195 kilometers (approximately 26 miles) under this pursued landmark time. It was not, however, an officially sanctioned world record, as it was not an "open race" of the IAAF. His time was 1 hour 59 minutes 40.2 seconds. Kipchoge ran in Vienna, Austria. It was an event specifically designed to help Kipchoge break the two hour barrier. & Kenyan runner Eliud Kipchoge has run a marathon in less than two hours.\\
\midrule
PG\&E stated it scheduled the blackouts in response to forecasts for high winds amid dry conditions. The aim is to reduce the risk of wildfires. Nearly 800 thousand customers were scheduled to be affected by the shutoffs which were expected to last through at least midday tomorrow. & Power has been turned off to millions of customers in California as part of a power shutoff plan.\\
\bottomrule
\end{tabular}
\caption{\label{table:xsum_samples} Example summaries from the XSum-tuned BART model on WikiNews articles. For clarity, only relevant excerpts of the source are shown. Summaries combine information from across the article and prior knowledge. 
}
\end{table*}
 
 We also evaluated performance on WMT16 Romanian-English, augmented with back-translation data from~\citet{sennrich2016}. We use a 6-layer transformer source encoder to map Romanian into a representation that BART is able to de-noise into English, following the approach introduced in \S\ref{sec:tuning-MT}.
 Experiment results are presented in Table~\ref{tab:MT_results}. We compare our results against a baseline Transformer architecture~\cite{vaswani:2017} with Transformer-large settings (the baseline row). We show the performance of both steps of our model in the fixed BART and tuned BART rows. For each row we experiment on the original WMT16 Romanian-English augmented with back-translation data. We use a beam width of 5 and a length penalty of $\alpha = 1$.
 Preliminary results suggested that our approach was less effective without back-translation data, and prone to overfitting---future work should explore additional regularization techniques.


\section{Qualitative Analysis}
\label{section:qualitative}




BART shows large improvements on summarization metrics, of up to 6 points over the prior state-of-the-art. To understand BART's performance beyond automated metrics, we analyse its generations qualitatively.

Table \ref{table:xsum_samples} shows example summaries generated by BART. Examples are taken from WikiNews articles published after the creation of the pre-training corpus, to eliminate the possibility of the events described being present in the model's training data. Following \citet{xsum}, we remove the first sentence of the article prior to summarizing it, so there is no easy extractive summary of the document. 

Unsurprisingly, model output is fluent and grammatical English. However, model output is also highly abstractive, with few phrases copied from the input. The output is also generally factually accurate, and integrates supporting evidence from across the input document with background knowledge (for example, correctly completing names, or inferring that PG\&E operates in California). In the first example, inferring that fish are protecting reefs from global warming requires non-trivial inference from the text. However, the claim that the work was published in Science is not supported by the source. 

These samples demonstrate that the BART pretraining has learned a strong combination of natural language understanding and generation.

\section{Related Work}

Early methods for pretraining were based on language models. GPT \cite{gpt} only models leftward context, which is problematic for some tasks. ELMo \cite{elmo} concatenates left-only and right-only representations, but does not pre-train interactions between these features.
\citet{gpt2} demonstrated that very large language models can act as unsupervised multitask models.

BERT \citep{bert} introduced masked language modelling, which allows pre-training to learn interactions between left and right context words. 
Recent work has shown that very strong performance can be achieved by training for longer \citep{roberta}, by tying parameters across layers \citep{albert}, and by masking spans instead of words \citep{spanbert}.
Predictions are not made auto-regressively, reducing the effectiveness of BERT for generation tasks.

UniLM \citep{unilm} fine-tunes BERT with an ensemble of masks, some of which allow only leftward context. Like BART, this allows UniLM to be used for both generative and discriminative tasks. A difference is that UniLM predictions are conditionally independent, whereas BART's are autoregressive. BART reduces the mismatch between pre-training and generation tasks, because the decoder is always trained on uncorrupted context. 

MASS \citep{mass} is perhaps the most similar model to BART. An input sequence where a contiguous span of tokens is masked is mapped to a sequence consisting of the missing tokens. MASS is less effective for discriminative tasks, because disjoint sets of tokens are fed into the encoder and decoder.

XL-Net \cite{xlnet} extends BERT by predicting masked tokens auto-regressively in a permuted order. This objective allows predictions to condition on both left and right context. In contrast, the BART decoder works left-to-right during pre-training, matching the setting during generation.

Several papers have explored using pre-trained representations to improve machine translation. The largest improvements have come from pre-training on both source and target languages \citep{mass,xlm}, but this requires pre-training on all languages of interest. Other work has shown that encoders can be improved using pre-trained representations \cite{edunov2019pre}, but gains in decoders are more limited. We show how BART can be used to improve machine translation decoders.

\section{Conclusions}
We introduced BART, a pre-training approach that learns to map corrupted documents to the original. 
BART achieves similar performance to RoBERTa on discriminative tasks, while achieving new state-of-the-art results on a number of text generation tasks. Future work should explore new methods for corrupting documents for pre-training, perhaps tailoring them to specific end tasks.

\clearpage


\bibliography{iclr2019_conference}
\bibliographystyle{iclr2019_conference}

\end{document}